%% file: rmn.tex
\pdfoutput=1
\documentclass[11pt,letterpaper]{article}
\usepackage{naaclhlt2016}
\usepackage{times}
\usepackage{latexsym}

\usepackage[usenames,dvipsnames]{color}
\usepackage{xcolor}
\usepackage{graphicx}
\usepackage{amsmath}
\usepackage{bm}
\usepackage{url}
\usepackage{setspace}

\usepackage{xspace}
\usepackage{amsfonts}

\newcommand{\R}{{\mathbb R}}
\newcommand{\trans}{\mathsf{T}}
\newcommand{\vt}[1]{\bm{\mathbf{#1}}}

\newcommand{\eg}{e.\,g.\xspace}
\newcommand{\vs}{vs.\xspace}

\usepackage{tabularx}
\usepackage{multirow}
\usepackage{booktabs}
\usepackage{caption}
\usepackage{subcaption}

\usepackage[linewidth=0.5pt]{mdframed}
\usepackage{tikz-dependency}
\newcommand{\gold}[1]{\textbf{#1}}
\newcommand{\predlstm}{\diamondsuit}
\newcommand{\predrmn}{\clubsuit}

\newcommand{\attw}{figs/sample_att.pdf}

\newcommand{\mb}{figs/mb.pdf}
\newcommand{\rmn}{figs/rmn_bw.pdf}
\newcommand{\sota}{69.2\%\xspace}

\newcommand{\tml}{+tM-l} % temporal memory linear
\newcommand{\bml}{--tM-l} % bag of words memory linear
\newcommand{\tmg}{+tM-g} % temporal memory gated
\newcommand{\bmg}{--tM-g} % bag of words memory gated
\newcommand \COMMENT[1]{}

\naaclfinalcopy % Uncomment this line for the final submission
\title{Recurrent Memory Networks for Language Modeling}

\author{Ke Tran \quad Arianna Bisazza \quad Christof Monz\\
	    Informatics Institute, University of Amsterdam\\
	    Science Park 904, 1098 XH Amsterdam, The Netherlands\\
	    {\tt \{m.k.tran,a.bisazza,c.monz\}@uva.nl}}

\date{}

\begin{document}
\maketitle

\begin{abstract}
Recurrent Neural Networks (RNNs) have obtained excellent result in many natural language processing (NLP) tasks. However, understanding and interpreting the source of this success remains a challenge. In this paper, we propose Recurrent Memory Network (RMN), a novel RNN architecture, that not only amplifies the power of RNN  but also facilitates our understanding of its internal functioning and allows us to discover underlying patterns in data. We demonstrate the power of RMN on language modeling and sentence completion tasks. On language modeling, RMN outperforms Long Short-Term Memory (LSTM) network on three large German, Italian, and English dataset. Additionally we perform in-depth analysis of various linguistic dimensions that RMN captures. On Sentence Completion Challenge, for which it is essential to capture sentence coherence, % to solve the task, 
our RMN obtains \sota accuracy, surpassing the previous state of the art by a large margin.\footnote{Our code and data are available  at \url{https://github.com/ketranm/RMN}}
\end{abstract}

\section{Introduction}

Recurrent Neural Networks (RNNs) \cite{elman:1990,mikolov:2010} are remarkably powerful models for sequential data. Long Short-Term Memory (LSTM) \cite{hochreiter:1997}, a specific architecture of RNN, has a track record of success in many natural language processing tasks such as language modeling \cite{jozefowicz:2015}, dependency parsing \cite{dyer:2015}, sentence compression \cite{filippova:2015}, and machine translation \cite{sutskever:2014}. 

Within the context of natural language processing, a common assumption is that LSTMs are able to capture certain linguistic phenomena. Evidence supporting this assumption mainly comes from evaluating LSTMs in downstream applications: \newcite{bowman:2015} carefully design two artificial datasets where sentences have explicit recursive structures. They show empirically that while processing the input linearly, LSTMs can implicitly exploit recursive structures of languages. \newcite{filippova:2015} find that using explicit syntactic features within LSTMs in their sentence compression model hurts the performance of overall system. They then hypothesize that a basic LSTM is powerful enough to capture syntactic aspects which are useful for compression.

To understand and explain which linguistic dimensions are captured by an LSTM is non-trivial. This is due to the fact that the sequences of input histories are compressed into several dense vectors by the LSTM's components whose purposes with respect to representing linguistic information is not evident.
To our knowledge, the only attempt to better understand the reasons of an LSTM's performance and limitations is the work of \newcite{karpathy:15} by means of visualization experiments and cell activation statistics in the context of character-level language modeling.
 
Our work is motivated by the difficulty in understanding and interpreting existing RNN architectures from a linguistic point of view. We propose Recurrent Memory Network (RMN), a novel RNN architecture that combines the strengths of both LSTM and Memory Network \cite{sukhbaatar:2015}. 
%RMN consists of two building blocks: an LSTM unit and a Memory Block. \AB{this sentence is redundant with previous} 
%The Memory Block takes the output of an LSTM as its input. Additionally, a Memory Block accesses the most recent input words and learns which words are relevant for making the next word prediction. 
% AB: TRYING TO SIMPLIFY:
In RMN, the Memory Block component---a variant of Memory Network---accesses the most recent input words and selectively attends to words that are relevant for predicting the next word given the current LSTM state. 
%The Memory Block achieves this goal by implementing an attention mechanism. 
By looking at the attention distribution over history words, our RMN allows us not only to interpret the results but also to discover underlying dependencies present in the data.

In this paper, we make the following contributions:
\begin{enumerate}
\item We propose a novel RNN architecture that complements LSTM in language modeling. We demonstrate that our RMN outperforms competitive LSTM baselines in terms of perplexity on three large German, Italian, and English datasets.
\item We perform an analysis along various linguistic dimensions that our model captures. This is possible only because the Memory Block allows us to look into its internal states and its explicit use of additional inputs at each time step.
\item  We show that, with a simple modification, our RMN can be successfully applied to NLP tasks other than language modeling. On the Sentence Completion Challenge \cite{zweig:2012}, our model achieves an impressive \sota accuracy, surpassing the previous state of the art 58.9\% by a large margin. % \AB{mention great sentence completion results :)}
\end{enumerate}

\section{Recurrent Neural Networks}\label{sec:lstm}
Recurrent Neural Networks (RNNs) have shown impressive performances on many sequential modeling tasks due to their ability to encode unbounded input histories. However, training simple RNNs is difficult because of the vanishing and exploding gradient problems \cite{bengio:1994,pascanu:2013}. A simple and effective solution for exploding gradients is gradient clipping proposed by \newcite{pascanu:2013}. To address  the more challenging problem of vanishing gradients, several variants of RNNs have been proposed. Among them, Long Short-Term Memory \cite{hochreiter:1997} and Gated Recurrent Unit \cite{cho:2014} are widely regarded as the most successful variants. In this work, we focus on LSTMs because they have been shown to outperform GRUs on language modeling tasks \cite{jozefowicz:2015}. In the following, we will detail the LSTM architecture used in this work.

\noindent \textbf{Long Short-Term Memory}

\noindent\emph{Notation}: Throughout this paper, we denote matrices, vectors, and scalars using bold uppercase (\eg, $\vt{W}$), bold lowercase (\eg, $\vt{b}$) and lowercase (\eg, $\alpha$) letters, respectively.

The LSTM used in this work is specified as follows:
\begin{align}
\vt{i}_t &= \text{sigm}(\vt{W}_{xi} \vt{x}_t + \vt{W}_{hi}\vt{h}_{t-1} + \vt{b}_i) \nonumber\\
\vt{j}_t &= \text{sigm}(\vt{W}_{xj} \vt{x}_t + \vt{W}_{hj}\vt{h}_{t-1} + \vt{b}_j) \nonumber\\
\vt{f}_t &= \text{sigm}(\vt{W}_{xf} \vt{x}_t + \vt{W}_{hf}\vt{h}_{t-1} + \vt{b}_f) \nonumber\\
\vt{o}_t &= \text{tanh}(\vt{W}_{xo} \vt{x}_t + \vt{W}_{ho}\vt{h}_{t-1} + \vt{b}_o) \nonumber\\
\vt{c}_t &= \vt{c}_{t-1} \odot \vt{f}_t + \vt{i}_t \odot \vt{j}_t \nonumber\\
\vt{h}_t &= \text{tanh}(\vt{c}_t) \odot \vt{o}_t \nonumber
\end{align}
where $\vt{x}_t$ is the input vector at time step $t$, $\vt{h}_{t-1}$ is the LSTM hidden state at the previous time step, $\vt{W}_{\ast}$ and $\vt{b}_{\ast}$ are weights and biases. The symbol $\odot$ denotes the Hadamard product or  element-wise multiplication. 
%%% CAMERA READY? %%%
%Thus the LSTM has an input gate $\vt{i}_t$, a forget gate $\vt{o}_t$, a cell input $\vt{j}_t$ and two types of hidden states, a cell state or \emph{slow} state $\vt{c}_t$ and a \emph{fast} state $\vt{h}_t$. While the design of the \emph{slow} state $\vt{c}_t$ is intended to avoid vanishing gradients by reparametrization, \KT{it is not immediately clear how the \emph{fast} state $\vt{h}_t$ assists LSTM in making complex decisions by maintaining relevant information over time.}

Despite the popularity of LSTM in sequential modeling, its design is not straightforward to justify and understanding why it works remains a challenge \cite{hermans:2013,chung:2014,greff:2015,jozefowicz:2015,karpathy:15}. 
There have been few recent attempts to understand the components of an LSTM from an empirical point of view: \newcite{greff:2015} carry out a large-scale experiment of eight LSTM variants. The results from their 5,400 experimental runs suggest that forget gates and output gates are the most critical components of LSTMs. \newcite{jozefowicz:2015} conduct and evaluate over ten thousand RNN architectures and find that the initialization of the forget gate bias is crucial to the LSTM's performance. While these findings are important to help choosing appropriate LSTM architectures, they do not shed light on what information is captured by the hidden states of an LSTM. 

%\todo{explain the problem explicitly (??)}
%\begin{figure}[htbp]
%\centering
%\includegraphics[scale=0.4]{\lstm}
%\caption{A vanila LSTM. \AB{is this figure ever used?} \KT{Christof thinks we should have this figure. I think we can omit it. I don't use this at the moment to see if we have enough space.}}
%\label{fig:lstm}
%\end{figure}

\newcite{bowman:2015} show that a vanilla LSTM, such as described above, performs reasonably well compared to a recursive neural network \cite{socher:2011} that explicitly exploits tree structures on two artificial datasets. They find that LSTMs can effectively exploit recursive structure in the artificial datasets. In contrast to these simple datasets containing a few logical operations in their experiments, natural languages exhibit highly complex patterns. The extent to which linguistic assumptions about syntactic structures and compositional semantics are reflected in LSTMs is rather poorly understood. Thus it is desirable to have a more principled mechanism allowing us to inspect recurrent architectures
from a linguistic perspective. In the following section, we propose such a mechanism.

\section{Recurrent Memory Network}
\noindent%
It has been demonstrated that RNNs can retain input information over a long period.
However, existing RNN architectures make it difficult to analyze what information is exactly retained at their hidden states at each time step, especially when the data has complex underlying structures, which is common in natural language. 
Motivated by this difficulty, we propose a novel RNN architecture called Recurrent Memory Network (RMN).  On linguistic data, the RMN allows us not only to qualify which linguistic information is preserved over time and why this is the case but also to discover dependencies within the data (Section~\ref{sec:analysis}). Our RMN consists of two components: an LSTM and a \emph{Memory Block} (MB) (Section~\ref{sec:memory}). The MB takes the hidden state of the LSTM and compares it to the most recent inputs % by a simple 
using an attention mechanism \cite{gregor:2015,bahdanau:2014,graves:2014}. 
Thus, analyzing the attention weights of a trained model can give us valuable insight into the information that is retained over time in the LSTM.

In the following, we describe in detail the MB architecture and the combination of the MB and the LSTM to form an RMN.

\subsection{Memory Block}\label{sec:memory}
%%%%CAMERA READY%%%%%
%At the core of our model is a Memory Block which allows us to qualify the contribution of each individual word in the history to the prediction of the next word. This is accomplished by explicitly comparing the similarity between each word and the hidden state of an LSTM, which encodes the whole history from the beginning of the sentence up to the current word. The degree of similarity reflects how much of the content of a word in the history is retained by the LSTM.
%%% END CAMERA READY%%%%%
\begin{figure}[htbp]
\centering
\includegraphics[scale=0.45]{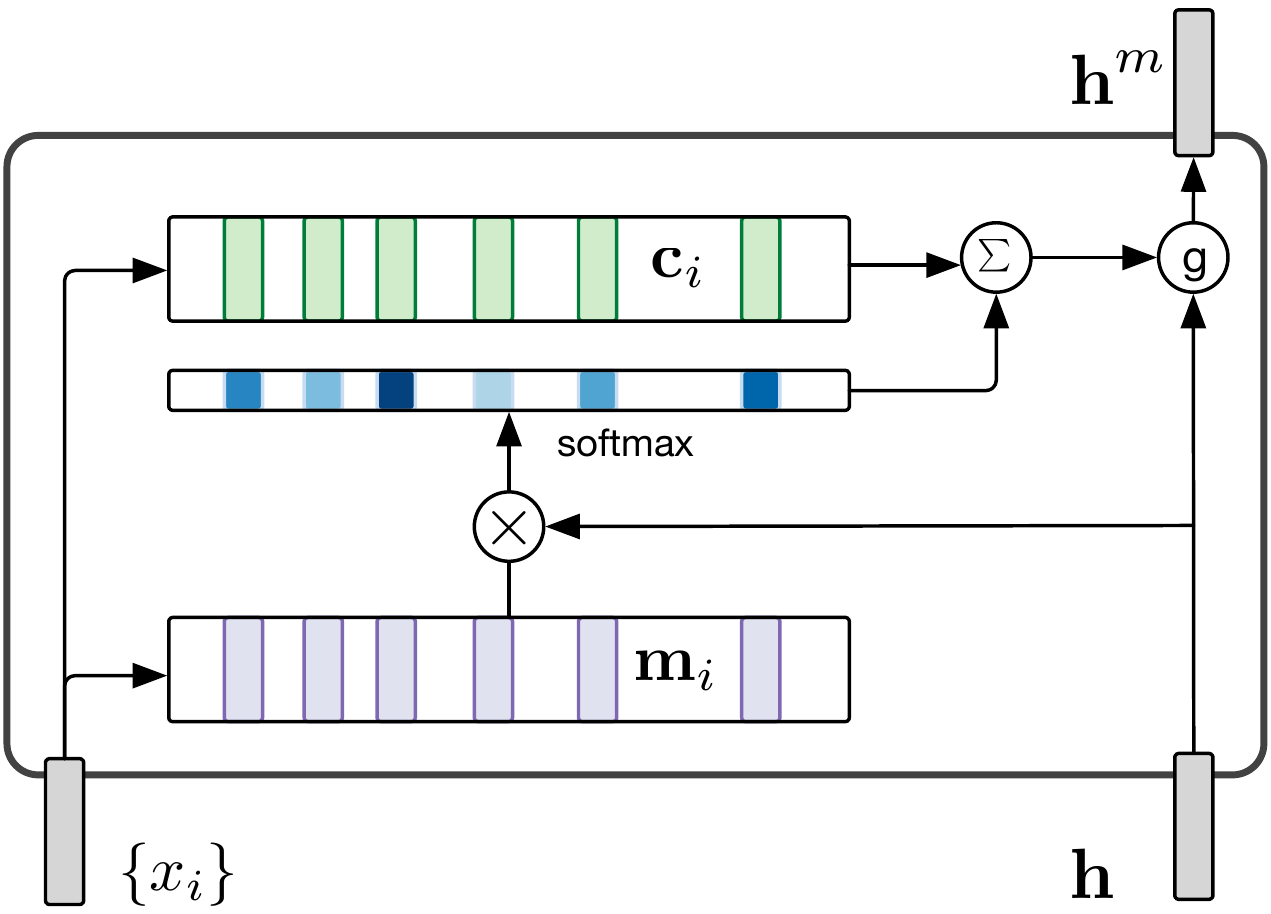}
\caption{A graphical representation of the MB.}
\label{fig:memblock}
\end{figure}
The Memory Block (Figure~\ref{fig:memblock}) is a variant of Memory Network \cite{sukhbaatar:2015} with one hop (or a single-layer Memory Network). At time step $t$, the MB receives two inputs: the hidden state $\vt{h}_t$ of the LSTM and a set $\{x_i\}$ of $n$ most recent words including the current word $x_t$. We refer to $n$ as the memory size. Internally, the MB consists of two lookup tables $\vt{M}$ and $\vt{C}$ of size $|V| \times d$, where $|V|$ is the size of the vocabulary. With a slight abuse of notation we denote $\vt{M}_i = \vt{M}(\{x_i\})$ and $\vt{C}_i = \vt{C}(\{x_i\})$ as $n \times d$ matrices where each row corresponds to an input memory embedding $\vt{m}_i$ and an output memory embedding $\vt{c}_i$ of each element of the set $\{x_i\}$.
We use the matrix $\vt{M}_i$ to compute an attention distribution over the set $\{x_i\}$:
\begin{equation}\label{eq:att}
\vt{p}_t = \text{softmax}(\vt{M}_i\vt{h}_t)
\end{equation}
When dealing with data that exhibits a strong temporal relationship, such as natural language, an additional temporal matrix $\vt{T} \in \R^{n\times d}$ can be used to bias attention with respect to the position of the data points. In this case, equation~\ref{eq:att} becomes
\begin{equation}\label{eq:att_temp}
\vt{p}_t = \text{softmax}\big((\vt{M}_i+\vt{T})\vt{h}_t\big)
\end{equation}
We then use the attention distribution $\vt{p}_t$ to compute a context vector representation of $\{x_i\}$:
\begin{equation}
\vt{s}_t = \vt{C}_i^\trans\vt{p}_t
\end{equation}
Finally, we combine the context vector $\vt{s}_t$ and the hidden state $\vt{h}_t$ by a function $g(\cdot)$ to obtain the output $\vt{h}_t^m$ of the MB. Instead of using a simple addition function $g(\vt{s}_t,\vt{h}_t) = \vt{s}_t+\vt{h}_t$  as in \newcite{sukhbaatar:2015}, we propose to use a gating unit that decides how much it should trust the hidden state $\vt{h}_t$ and context $\vt{s}_t$ at time step $t$. Our gating unit is a form of Gated Recurrent Unit \cite{cho:2014,chung:2014}:
\begin{align}
\vt{z}_t &= \text{sigm}(\vt{W}_{sz}\vt{s}_t + \vt{U}_{hz}\vt{h}_t) \\
\vt{r}_t &= \text{sigm}(\vt{W}_{sr}\vt{s}_t + \vt{U}_{hr}\vt{h}_t) \\
\tilde{\vt{h}}_t &= \text{tanh}(\vt{W}\vt{s}_t + \vt{U}(\vt{r}_t \odot \vt{h}_t)) \\
\vt{h}_t^m &= (\vt{1}-\vt{z}_t) \odot \vt{h}_t + \vt{z}_t\odot\tilde{\vt{h}}_t 
\end{align}
where $\vt{z}_t$ is an update gate, $\vt{r}_t$ is a reset gate.

The choice of the composition function $g(\cdot)$ is crucial for the MB especially when one of its input comes from the LSTM. The simple addition function might overwrite the information within the LSTM's hidden state and therefore prevent the MB from keeping track of information in the distant past. The gating function, on the other hand, can control the degree of information that flows from the LSTM to the MB's output.

\subsection{RMN Architectures}

As explained above, our proposed MB receives the hidden state of the LSTM as one of its input. This leads to an intuitive combination of the two units by stacking the MB on top of the LSTM. We call this architecture Recurrent-Memory (RM). 
The RM architecture, however, does not allow interaction between Memory Blocks at different time steps. 
To enable this interaction we can stack one more LSTM layer on top of the RM. We call this architecture Recurrent-Memory-Recurrent (RMR).

\begin{figure}[htbp]
\centering
\includegraphics[scale=0.35]{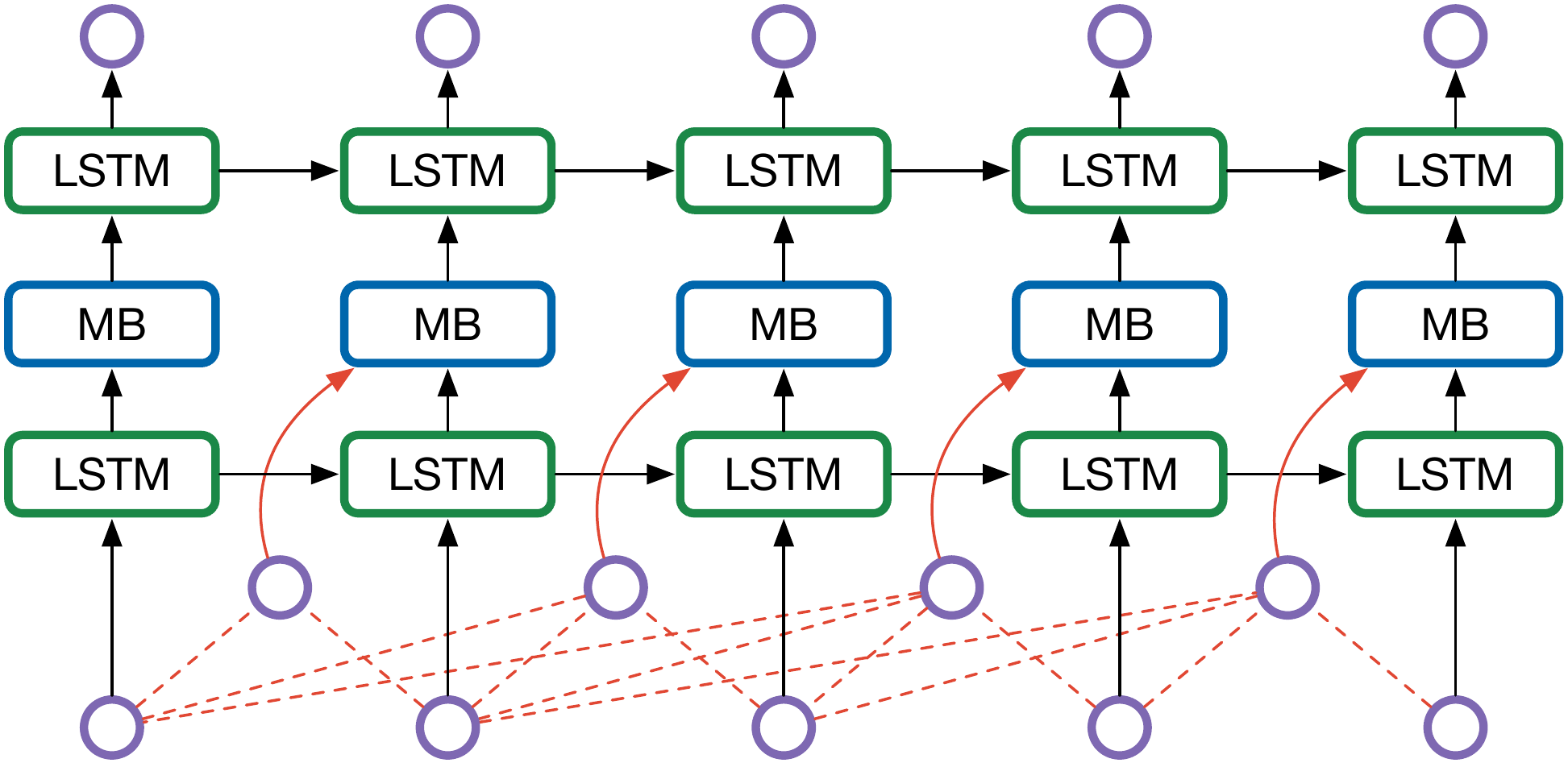}
\caption{A graphical illustration of an unfolded RMR with memory size 4. Dashed line indicates concatenation. The MB takes the output of the bottom LSTM layer and the 4-word history as its input. The output of the MB is then passed to the second LSTM layer on top. There is no direct connection between MBs of different time steps. The last LSTM layer carries the MB's outputs recurrently.}
\label{fig:rmr}
\end{figure}

\section{Language Model Experiments}\label{sec:ppl}
Language models play a crucial role in many NLP applications such as machine translation and speech recognition. Language modeling also serves as a standard test bed for newly proposed models \cite{sukhbaatar:2015,kalchbrenner:2015}. We conjecture that, by explicitly accessing history words, RMNs will offer better predictive power than the existing recurrent architectures. We therefore evaluate our RMN architectures against state-of-the-art LSTMs in terms of perplexity.

\subsection{Data}
We evaluate our models on three languages: English, German, and Italian. We are especially interested in German and Italian because of their larger vocabularies and complex agreement patterns. Table~\ref{tb:data} summarizes the data used in our experiments. 

\input{table-data.tex}

The training data correspond to approximately 1M sentences in each language.
For English, we use all the News Commentary data (8M tokens) and 18M tokens from News Crawl 2014 for training. Development and test data are randomly drawn from the concatenation of the WMT 2009-2014 test sets \cite{wmt:2015}.
For German, we use the first 6M tokens from the News Commentary data and 16M tokens from News Crawl 2014 for training. For development and test data we use the remaining part of the News Commentary data concatenated with the WMT 2009-2014 test sets.
Finally, for Italian, we use a selection of 29M tokens from the PAIS\`{A} corpus \cite{paisa-corpus:14}, mainly including Wikipedia pages and, to a minor extent, Wikibooks and Wikinews documents.
For development and test we randomly draw documents from the same corpus.

\subsection{Setup}\label{sec:setup}

Our baselines are a 5-gram language model with Kneser-Ney smoothing, a Memory Network (MemN) \cite{sukhbaatar:2015}, a vanilla single-layer LSTM, and two stacked LSTMs with two and three layers respectively. 
N-gram models have been used intensively in many applications for their excellent performance and fast training. \newcite{chen:2015} show that n-gram model outperforms  a popular feed-forward language model \cite{bengio:2003} on a one billion word benchmark \cite{chelba:2013}. While taking longer time to train, RNNs have been proven superior to n-gram models.

We compare these baselines with our two model architectures: RMR and RM. 
For each of our models, we consider two settings: with or without temporal matrix (+tM or --tM), and linear \vs gating composition function. In total, we experiment with eight RMN variants.

For all neural network models, we set the dimension of word embeddings, the LSTM hidden states, its gates, the memory input, and output embeddings to 128. The memory size is set to 15. The bias of the LSTM's forget gate is initialized to 1 \cite{jozefowicz:2015} while all other parameters are initialized uniformly in $(-0.05,0.05)$.
%The initial learning rate is set to 1 and is halved at each epoch after the forth. All models are trained for 15 epochs with standard stochastic gradient descent (SGD). During training, we rescale the gradients whenever their norm is greater than 5 as suggested by \newcite{pascanu:2013}.
The initial learning rate is set to 1 and is halved at each epoch after the forth epoch. All models are trained for 15 epochs with standard stochastic gradient descent (SGD). During training, we rescale the gradients whenever their norm is greater than 5  \cite{pascanu:2013}.

Sentences with the same length are grouped into buckets. Then, mini-batches of 20 sentences are drawn from each bucket. We do not use truncated back-propagation through time, instead gradients are fully back-propagated from the end of each sentence to its beginning. When feeding in a new mini-batch, the hidden states of LSTMs are reset to zeros, which ensures that the data is properly modeled at the sentence level. 
%%%%CAMERA READY%%%%%
%\AB{These measures, sometimes neglected in the literature, turn out to be essential to build competitive language modeling baselines. (mention perplexity differences in preliminary experiments?)}
%%%%END CAMERA READY%%%%%
%For our RMN models, we do not use padding when the number of history words is less than the memory size. Instead, at time step $t < n$, we use a slice $\vt{T}_i = \vt{T}[1:t]\in \R^{t\times d}$ of the temporal matrix  $\vt{T}\in\R^{n\times d}$. When $t\ge n$, we set $\vt{T}_i = \vt{T}$.
For our RMN models, instead of using padding, at time step $t < n$, we use a slice $\vt{T}[1:t]\in \R^{t\times d}$ of the temporal matrix  $\vt{T}\in\R^{n\times d}$. %When $t\ge n$, we set $\vt{T}_i = \vt{T}$.

\subsection{Results}
\input{table-ppl.tex}
Perplexities on the test data are given in Table~\ref{tb:ppl}.
All RMN variants largely outperform $n$-gram and MemN models,
and most RMN variants also outperform the competitive LSTM baselines. 
The best results overall are obtained by RM with temporal matrix and gating composition (+tM-g).

%The results also reflect that gating composition is essential to our models. 
Our results agree with the hypothesis of mitigating prediction error by explicitly using the last $n$ words in RNNs \cite{karpathy:15}. We further observe that using a temporal matrix always benefits the RM architectures. This can be explained by seeing the RM as a principled way to combine an LSTM and a neural $n$-gram model. 
By contrast, RMR works better without temporal matrix but its overall performance is not as good as RM.
%The perplexity, however, increases in RMR when temporal matrix is applied. Moreover RMR, with its additional LSTM on top, performs worse than RM. 
This suggests that we need a better mechanism to address the interaction between MBs, which we leave to future work.
Finally, the proposed gating composition function outperforms the linear one in most cases.  

For historical reasons, we also run a stacked three-layer LSTM and a RM(+tM-g) on the much smaller Penn Treebank dataset \cite{marcus:1993} with the same setting described above. The respective perplexities are 126.1 and 123.5. %While the gain is smaller on this data set
%We suspect the smaller gain is due to the very limited size of this corpus (around 4\% of our data sets).

\section{Attention Analysis}\label{sec:analysis}
The goal of our RMN design is twofold: (i) to obtain better predictive power and (ii) to facilitate understanding of the model and discover patterns in data. In Section~\ref{sec:ppl}, we have validated the predictive power of the RMN and below we investigate the source of this performance based on linguistic assumptions of word co-occurrences and dependency structures.

\subsection{Positional and lexical analysis}
\label{sec:gen-analysis}

\begin{figure}[t]\centering
{\includegraphics[width=.9\linewidth]{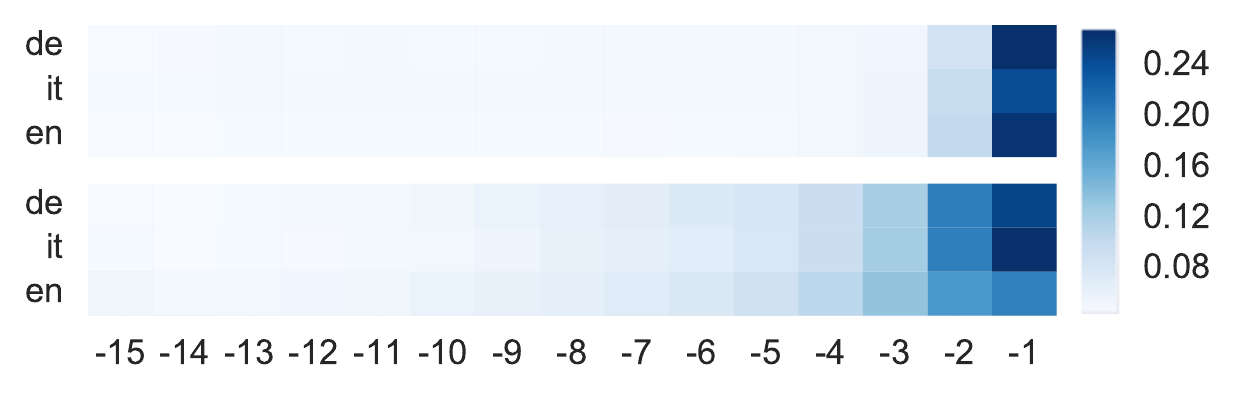}}
\caption{Average attention per position of RMN history. Top: RMR(\bmg), bottom:  RM(\tmg). Rightmost positions represent most recent history.} % Darker color means higher weight. }
\label{fig:agv_att}
\end{figure}

As a first step towards understanding RMN, we look at the average attention weights of each history word position in the MB of our two best model variants (Figure~\ref{fig:agv_att}).
One can see that the attention mass tends to concentrate at the rightmost position (the current word) and decreases when moving further to the left (less recent words). This is not surprising since the success of $n$-gram language models has demonstrated that the most recent words provide important information for predicting the next word. Between the two variants, the RM average attention mass is less concentrated to the right. This can be explained by the absence of an LSTM layer on top, meaning that the MB in the RM architecture has to pay more attention to the more distant words in the past. The remaining analyses described below are performed on the RM(\tmg) architecture as this yields the best perplexity results overall.

\begin{figure}[bp]
\centering
\includegraphics[scale=0.50]{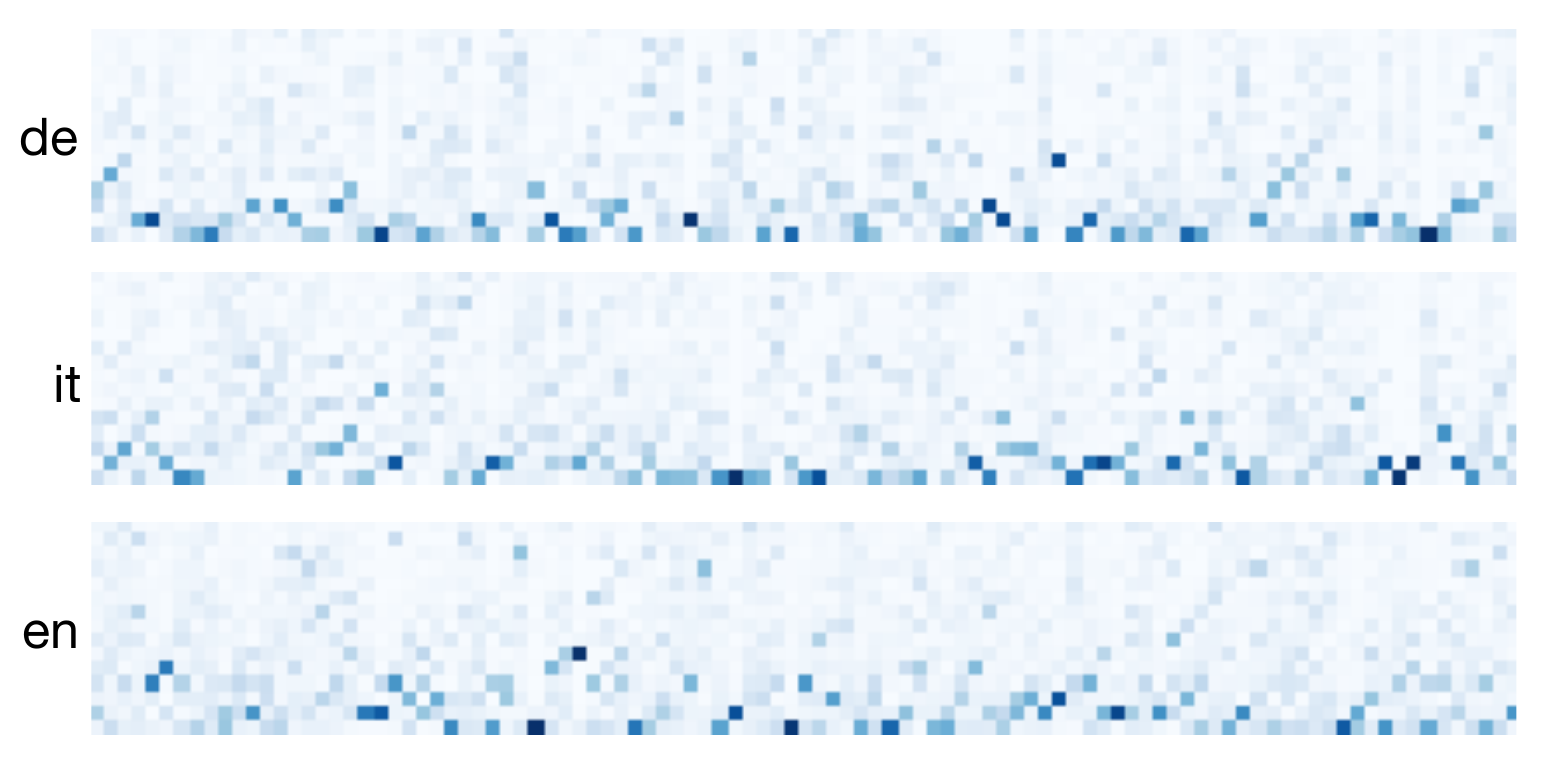}
\caption{Attention visualization of 100 word samples. Bottom positions in each plot represent most recent history. Darker color means higher weight. }
\label{fig:sample_att}
\end{figure}
% COMMENT OUT 
%\input{table-attend-examples.tex}
\begin{figure*}
\centering
   {\includegraphics[width=.95\linewidth]{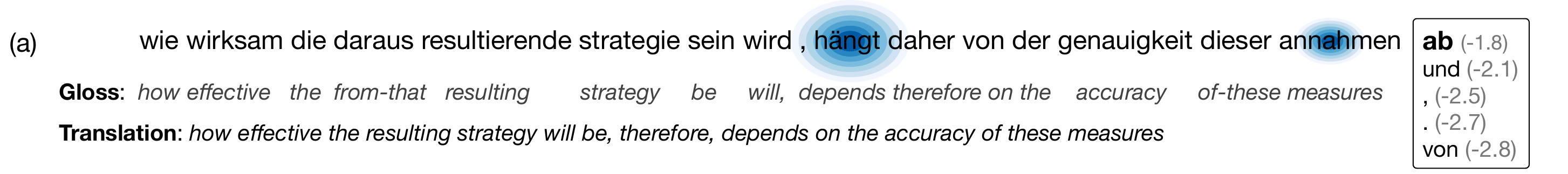} \phantomsubcaption\label{fig:example_sv} }\\[-.2ex]
   {\includegraphics[width=.95\linewidth]{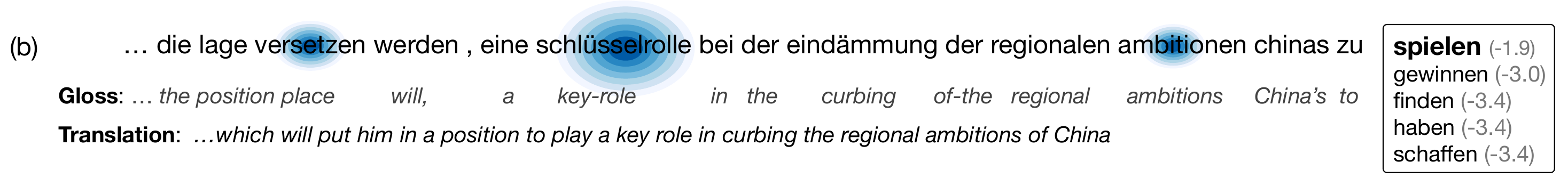} \phantomsubcaption\label{fig:example_fe} }\\[-.2ex]
   {\includegraphics[width=.95\linewidth]{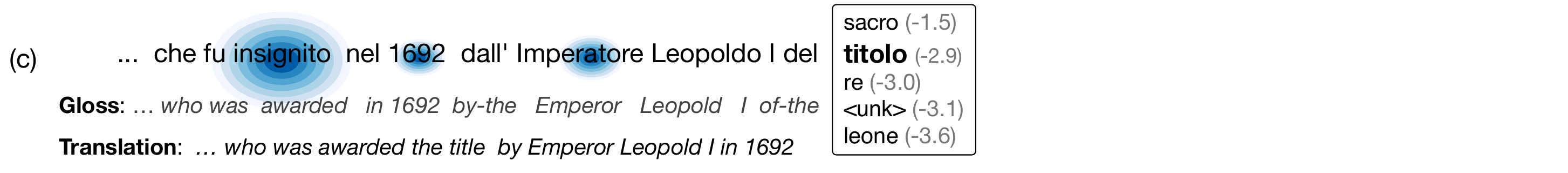} \phantomsubcaption\label{fig:example_it} }\\ \vspace{1mm}
\caption{Examples of distant memory positions attended by RMN. The resulting top five word predictions are shown with the respective log-probabilities.
% before predicting the last word. 
The correct choice (in bold) was ranked first in sentences (a,b) and second in (c).}
\label{fig:att-examples}
\end{figure*}
% sent1 (de): rank of gold word: 1
% sent2 (de): rank of gold word: 1
% sent3 (it): rank of gold word: 2

Beyond average attention weights, we are interested in those cases where attention focuses on distant positions. To this end, we randomly sample 100 words from test data and visualize attention distributions over the last 15 words. Figure~\ref{fig:sample_att} shows the attention distributions for random samples of German and Italian. Again, in many cases attention weights concentrate around the last word (bottom row). However, we observe that many long distance words also receive noticeable attention mass. Interestingly, for many predicted words, attention is distributed evenly over memory positions, possibly indicating cases where the LSTM state already contains enough information to predict the next word.

%%%%CAMERA READY%%%%%
%We are mainly interested in long distance dependencies between words and we hypothesize that our RMN captures some sort of co-occurrence frequency. We run our RM(\tmg)  model on 200,000 German sentences and select those pairs of (\emph{predicted word,} \emph{most attended word}) where the attention mass in the Memory Block concentrates at a position more than six words to the left.
%\AB{Give an example of pair?}
%Here, we only focus on content words, i.e., ignore high-frequency or stop-words. 
%Figure~\ref{fig:distfreq} depicts the relationship between attention distance and co-occurrence frequency.
%
%\begin{figure}[htbp]
%\centering
%\includegraphics[scale=0.4]{figs/dist_freq.pdf}
%\caption{Distance \vs co-occurrence. For visualization purpose, we removed those pairs with frequency is greater than 1,500.}
%\label{fig:distfreq}
%\end{figure}
%
%It is clear that our RMN learns co-occurrence implicitly. \KT{Previous work has studied this property of LSTM by analyzing a simple case of closing parenthesis and brace \cite{hermans:2013,karpathy:15}. RMN, on other hand, allows us to discover more interesting dependencies in the data TABLE OF INTERESTING CASES NEEDED HERE. To confirm our intuition that RMN keeps track of long distance dependencies, we compute the mean of co-occurrence frequencies per memory location (Table~\ref{tb:distfreq}) and find that there is no correlation between frequency and memory location.}
%%%%END CAMERA READY%%%%%

To explain the long-distance dependencies, we first hypothesize that our RMN mostly memorizes frequent co-occurrences. 
We run the RM(\tmg)  model on the German development and test sentences, and select those pairs of (\emph{most-attended-word, word-to-predict}) where the MB's attention concentrates on a word more than six positions to the left. %Then we compute the mean of co-occurrence frequencies seen in training data per memory location (Table~\ref{tb:distfreq}). This shows that our RMN learns co-occurrence implicitly. Moreover, we find that there is no correlation between frequency and memory location. 
Then, for each set of pairs with equal distance, we compute the mean frequency of corresponding co-occurrences seen in the training data (Table~\ref{tb:distfreq}). 
The lack of correlation between frequency and memory location suggests that RMN does more than simply memorizing frequent co-occurrences.

\input{table-freq.tex}

Previous work \cite{hermans:2013,karpathy:15} studied this property of LSTMs by analyzing simple cases of closing brackets. By contrast RMN allows us to discover more interesting dependencies in the data.
We manually inspect those high-frequency pairs to see whether they display certain linguistic phenomena. % or they just happen by chance. 
We observe that RMN captures, for example, \emph{separable verbs} and \emph{fixed expressions} in German. %German verbs can have separable prefixes.
Separable verbs are frequent in German: they typically consist of preposition+verb constructions, such \emph{ab+h\"angen} (`to depend') or \emph{aus+schlie{\ss}en} (`to exclude'), and can be spelled together (\emph{abh\"angen}) or apart as in `\emph{h\"angen von der Situation ab}' (`depend on the situation'), depending on the grammatical construction. 
Figure~\ref{fig:example_sv} shows a long-dependency example for the separable verb \emph{abh\"angen (to depend)}.
When predicting the verb's particle \emph{ab}, the model correctly attends to the verb's core \emph{h\"angt} occurring seven words to the left.
Figure~\ref{fig:example_fe} and \ref{fig:example_it} show fixed expression examples from German and Italian, respectively: \emph{schl\"usselrolle ... spielen (play a key role)} and \emph{insignito ... titolo (awarded title)}. Here too, the model correctly attends to the key word despite its long distance from the word to predict. 

Other interesting examples found by the RMN in the test data include:
\input{table-examples.tex}
\subsection{Syntactic analysis}

It has been conjectured that RNNs, and LSTMs in particular, model text so well because they capture syntactic structure implicitly. Unfortunately this has been hard to prove, but with our RMN model we can get closer to answering this important question.

\begin{figure}[ht]
\begin{subfigure}[b]{\columnwidth}
%\caption{German:}
\hspace{-1.8mm}
\includegraphics[scale=.555]{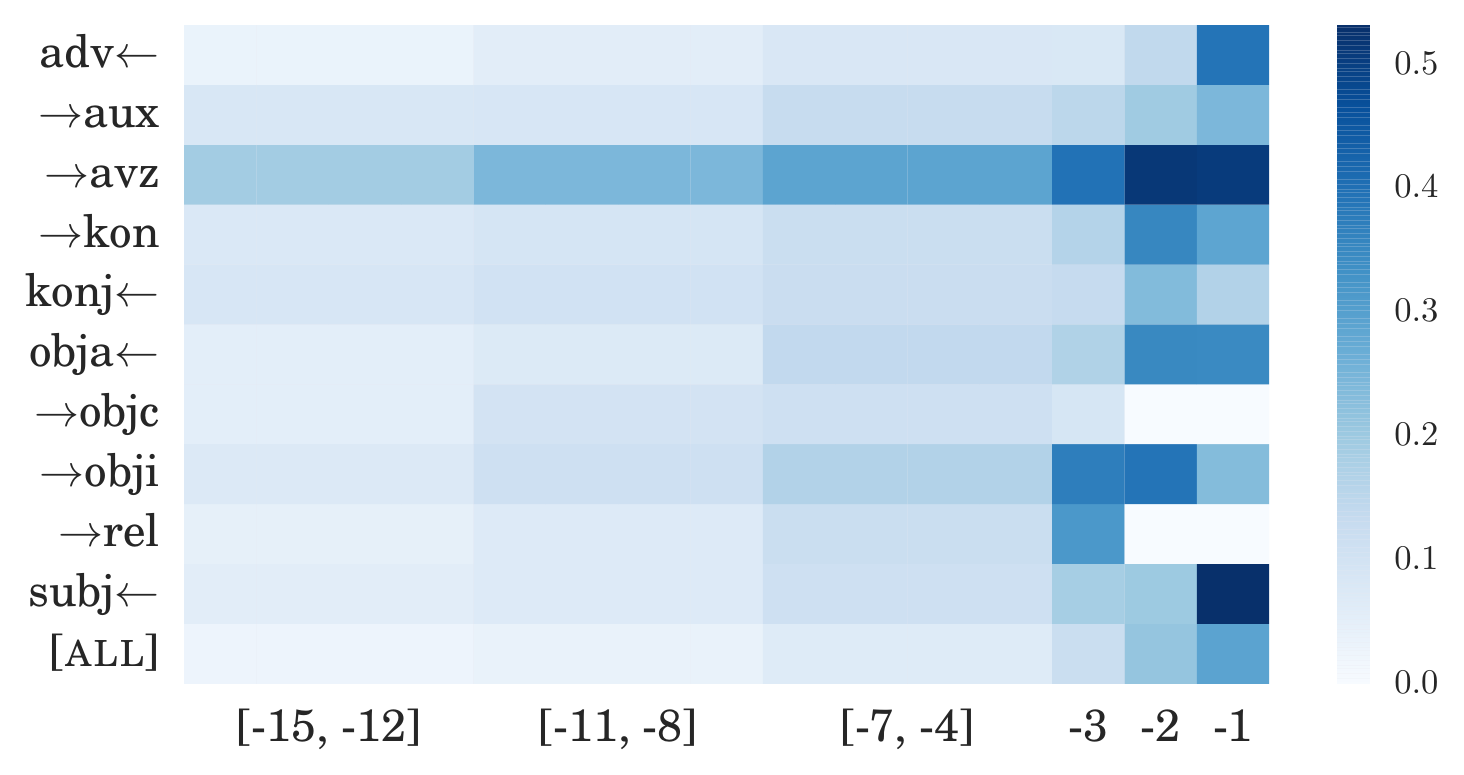}
\end{subfigure}
\begin{subfigure}[b]{\columnwidth}
%\caption{Italian:}
\hspace{-2.8mm}
\includegraphics[scale=.555]{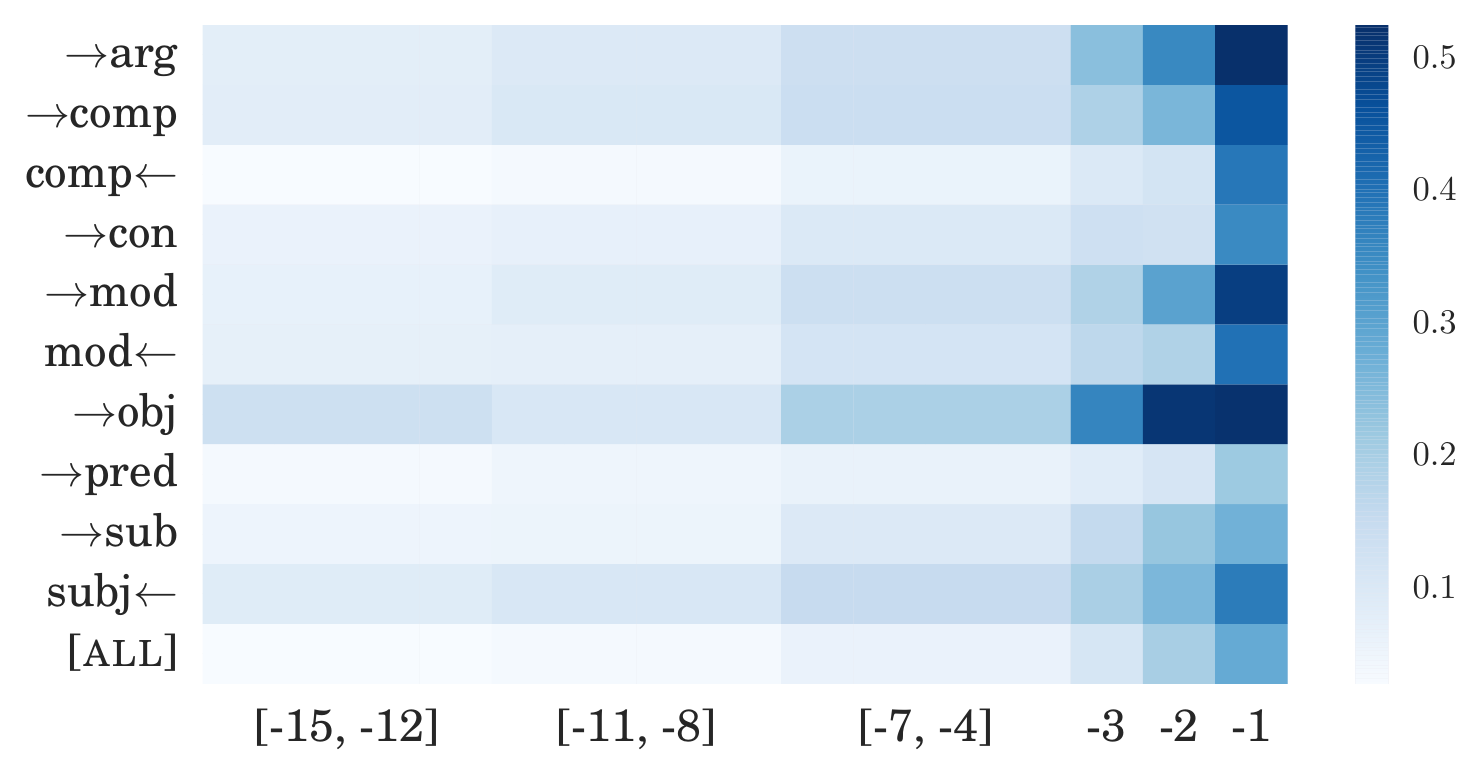}
\end{subfigure}
\caption{Average attention weights per position, broken down by dependency relation type+direction between the attended word and the word to predict. Top: German. Bottom: Italian. More distant positions are binned.
}
\label{fig:dep-att-it}
\end{figure}

We produce dependency parses for our test sets using \cite{parzu-de:13} for German and \cite{parser-it:09} for Italian.
Next we look at how much attention mass is concentrated by the RM(\tmg) model on different dependency types.
Figure~\ref{fig:dep-att-it} shows, for each language, a selection of ten dependency types that are often long-distance.%
\footnote{%Selected dependencies: \textit{arg}: argument, \textit{comp}: complement, copulative \textit{con}: conjunction,  \textit{mod}: modifier (adjectival, adverbial, nominal or clausal), \textit{obj}: direct object, \textit{pred}: predicative complement, \textit{sub}: subordinative conjunction, \textit{subj}: subject.
The full plots are available at \url{https://github.com/ketranm/RMN}.
The German and Italian tag sets are explained in \cite{depset-it:14} and \cite{depset-de:06} %\url{https://github.com/rsennrich/ParZu/blob/master/LABELS.md} 
respectively.}
Dependency direction is marked by an arrow: e.g. \textit{$\rightarrow$mod} means that the word to predict is a modifier of the attended word, while \textit{mod$\leftarrow$} means that the attended word is a modifier of the word to predict.\footnote{Some dependency directions, like \textit{obj$\leftarrow$} in Italian, are almost never observed due to order constraints of the language.}
White cells denote combinations of position and dependency type that were not present in the test data.

While in most of the cases closest positions are attended the most, we can see that some dependency types also receive %noticeable, if not uniform, attention on the long-distance positions.
noticeably more attention than the average (\textsc{all}) on the long-distance positions.
In German, this is mostly visible for the head of separable verb particles ($\rightarrow$\textit{avz}), which nicely supports our observations in the lexical analysis (Section~\ref{sec:gen-analysis}).
Other attended dependencies include: 
auxiliary verbs ($\rightarrow$\textit{aux}) when predicting the second element of a complex tense (\emph{hat\,\ldots gesagt / has said});
%\footnote{For space constraints we only provide very simple examples.};
subordinating conjunctions (\textit{konj}$\leftarrow$) when predicting the clause-final inflected verb
(\emph{\underline{dass} sie sagen \underline{sollten} / \underline{that} they \underline{should} say});
control verbs ($\rightarrow$\textit{obji}) when predicting the infinitive verb 
(\emph{\underline{versucht} ihr zu \underline{helfen} / \underline{tries} to \underline{help} her}).
Out of the Italian dependency types selected for their frequent long-distance occurrences (bottom of Figure~\ref{fig:dep-att-it}),
the most attended are argument heads ($\rightarrow$\textit{arg}), complement heads ($\rightarrow$\textit{comp}), object heads ($\rightarrow$\textit{obj}) and subjects (\textit{subj}$\leftarrow$). This suggests that RMN is mainly capturing predicate argument structure in Italian.
Notice that syntactic annotation is never used to train the model, but only to analyze its predictions.

%\begin{figure}[t]
%\centering
%\includegraphics[scale=.52]{figs/syn.pdf}
%\caption{Average attention weights per position, breakdown by dependency relation between the predicted word and the attended word, in German (top) and Italian (bottom). Distant positions are binned as follows: -15 to -12, -11 to -8, -7 to -4.
%}
%\label{fig:dep-att-it}
%\end{figure}

%\begin{figure}[t]
%\centering
%\hspace{1,5mm}\includegraphics[scale=.6]{figs/plot_select_att_1_de.pdf}
%%\includegraphics[scale=.619]{figs/plot_select_att_1_it.pdf}
%\includegraphics[scale=.55]{figs/syn_it.pdf}
%\caption{Average attention weights per position, breakdown by dependency relation between the predicted word and the attended word, in German (top) and Italian (bottom). Distant positions are binned as follows: -15 to -12, -11 to -8, -7 to -4.
%}
%\label{fig:dep-att-it}
%\end{figure}

We can also use RMN to discover which complex dependency paths are important for word prediction.
To mention just a few examples, high attention on the German path \textit{\small[\normalsize subj$\leftarrow$,$\rightarrow$kon,$\rightarrow$cj\small]} indicates that the model captures morphological agreement between coordinate clauses in non-trivial constructions of the kind: 
\textit{spielen die \underline{Kinder} im Garten und \underline{singen} /
the \underline{children} play in the garden and \underline{sing}}.
In Italian, high attention on the path \textit{\small[\normalsize$\rightarrow$obj,$\rightarrow$comp,$\rightarrow$prep\small]}
%suggests that RMN semantic dependency between 
denotes cases where the semantic relatedness between a verb and its object does not stop at the object's head, but percolates down to a prepositional phrase attached to it
(\textit{\underline{pass\`{o}} buona parte della sua \underline{vita} / \underline{spent} a large part of his \underline{life}}).
Interestingly, both local n-gram context and immediate dependency context would have missed these relations.

While much remains to be explored, our analysis shows that RMN discovers patterns far more complex than pairs of opening and closing brackets, 
and suggests that the network's hidden state captures to a large extent the underlying structure of text. 
%and discovers patterns that are far more complex than pairs of opening and closing parentheses.

%\AB{Future work: further investigate agreement phenomena (and more complex dependency paths?)}

\section{Sentence Completion Challenge}

The Microsoft Research Sentence Completion Challenge \cite{zweig:2012} has recently become a test bed for advancing statistical language modeling. We choose this task to demonstrate the effectiveness of our RMN in capturing sentence coherence. The test set consists of 1,040 sentences selected from five Sherlock Holmes novels by Conan Doyle. For each sentence, a content word is removed %and five candidates of the missing word are provided. 
and the task is to identify the correct missing word among five given candidates.
%chosen to be syntactically valid and compatible with the local context.
The task is carefully designed to be non-solvable for local language models such as $n$-gram models. The best reported result is 58.9\% accuracy \cite{mikolov:2013}\footnote{The authors use a weighted combination of skip-ngram and RNN without giving any technical details.} which is far below human accuracy of 91\% \cite{zweig:2012}.

As baseline we use a stacked three-layer LSTM. 
Our models are two variants of RM(+tM-g), each consisting of three LSTM layers followed by a MB.
The first variant (unidirectional-RM) uses $n$ words preceding the word to predict,
the second (bidirectional-RM) uses the $n$ words preceding \textit{and} the $n$ words following the word to predict, as MB input.
%(i) a RM that uses $n$ preceding words of the predicted word, and (ii) a RM that uses $n$ preceding words and $n$ following words of the predicted words as Memory Block input. The two variants consist of three layerfs LSTM following by a Memory Block. 
%For the sake of clarity, we refer to the former as unidirectional-RM and the later as bidirectional-RM. 
We include bidirectional-RM in the experiments to show the flexibility of utilizing future context in RMN. 

\input{fig-examples.tex}

We train all models on the standard training data of the challenge, which consists of 522 novels from Project Gutenberg, preprocessed similarly to \cite{mnih:2013}. After sentence splitting, tokenization and lowercasing, we randomly select 19,000 sentences for validation. Training and validation sets include 47M and 190K tokens respectively. The vocabulary size is about 64,000.

We initialize and train all the networks as described in Section~\ref{sec:setup}.
Moreover, for regularization, we place dropout \cite{srivastava:2014} after each LSTM layer as suggested in \cite{pham:2013}. The dropout rate is set to 0.3 in all the experiments. 

\input{table-acc.tex}

Table~\ref{tb:acc} summarizes the results. 
%%%% CAMERA READY %%%%
%In addition to the results of our three models, we also cite the accuracy of a dependency recurrent neural network (depRNN) reported in \cite{mirowski:2015}.  Essentially, given a dependency tree, depRNN is a simple RNN that takes dependency paths from leaf nodes to the root node as its sequential inputs. DepRNN, therefore, makes explicit use of syntactic information. As can be seen in table~\ref{tb:acc}, without  explicit knowledge of tree structure, LSTM model outperforms depRNN by more than 3 points. 
%%%%END CAMERA READY %%%%
It is worth to mention that our LSTM baseline %the second best published result (55.5\%) of an inverse log-bilinear model \cite{mnih:2013}, 
outperforms a dependency RNN making explicit use of syntactic information \cite{mirowski:2015}
and performs on par with the best published result \cite{mikolov:2013}.
Our unidirectional-RM sets a new state of the art for the Sentence Completion Challenge with \sota accuracy. Under the same setting of  $d$ we observe that using bidirectional context does not bring additional advantage to the model. \newcite{mnih:2013} also report a similar observation. We believe that RMN may achieve further improvements with hyper-parameter optimization. %\cite{maclaurin:2015}.

Figure~\ref{fig:sentcomp-examples} shows some examples where our best RMN beats the already very competitive LSTM baseline, or where both models fail.
We can see that in some sentences the necessary clues to predict the correct word occur only to its \textit{right}. 
While this seems to conflict with the worse result obtained by the bidirectional-RM, it is important to realize that prediction corresponds to the whole sentence probability.
Therefore a badly chosen word can have a negative effect on the score of future words. This appears to be particularly true for the RMN due to its ability to directly access (distant) words in the history.
The better performance of unidirectional versus bidirectional-RM may indicate that the attention in the memory block can be distributed reliably only on words that have been already seen and summarized by the current LSTM state.
%cannot be distributed reliably on future words because the current LSTM state 
In future work, we may investigate whether different ways to combine two RMNs running in opposite directions further improve accuracy on this challenging task.

\section{Conclusion}
We have proposed the Recurrent Memory Network (RMN), a novel recurrent architecture for language modeling. Our RMN outperforms LSTMs in terms of perplexity on three large dataset and allows us to analyze its behavior from a linguistic perspective. 
We find that RMNs learn important co-occurrences regardless of their distance. 
Even more interestingly, our RMN implicitly captures certain dependency types that are important for word prediction, despite being trained without any syntactic information. 
Finally RMNs obtain excellent performance at modeling sentence coherence,
setting a new state of the art on the challenging sentence completion task.

\section*{Acknowledgments}

This research was funded in part by the Netherlands Organization for Scientific Research (NWO) under project numbers 639.022.213 and 612.001.218.

\bibliography{ke-biblio,naaclhlt2016}
\bibliographystyle{naaclhlt2016}

\end{document}

%% file: table-data.tex
\begin{table}[ht]\centering
\begin{tabular}{@{}l r r r r r r@{}}
\textbf{Lang} & \multicolumn{1}{c}{Train} & \multicolumn{1}{c}{Dev} & \multicolumn{1}{c}{Test} & \phantom{ab}& \multicolumn{1}{c}{$|s|$} &\multicolumn{1}{c}{$|V|$}\\
%\multicolumn{3}{c}{\#Sents / \#Toks} & $|V|$\\
%\cmidrule{2-4}
\toprule
En & 26M & 223K & 228K & & 26 &77K\\
De & 22M & 202K & 203K & & 22 & 111K\\
%De & 22M & 66K & 68K & & 22 & 111K\\
It & 29M & 207K & 214K & & 29 &104K\\
\bottomrule
\end{tabular}
\caption{Data statistics. $|s|$ denotes the average sentence length and $|V|$ the vocabulary size.} 
\label{tb:data}
\end{table}

%% file: table-ppl.tex
\begin{table}[ht]\centering
\begin{tabular}{@{}l l r r r@{}}
\bf{Model} & & \multicolumn{1}{c}{\bf{De}} & \multicolumn{1}{c}{\bf{It}} & \multicolumn{1}{c}{\bf{En}}\\
\toprule
 5-gram & -- & 225.8 & 167.5 & 219.0 \\
 \midrule
 MemN & 1 layer & 169.3 & 127.5 & 188.2\\
 \midrule
\multirow{3}{*}{LSTM}
 & 1 layer & 135.8 & 108.0 & 145.1\\
 & 2 layers & 128.6 & 105.9 & 139.7\\
 & 3 layers & 125.1 & 106.5 & 136.6\\
\midrule
\multirow{4}{*}{RMR}
 & \tml & 127.5 & 109.9 & 133.3\\
 & \bml & 126.4 & 106.1 & 134.5\\
 \cmidrule{2-5}
 & \tmg & 126.2 & 99.5 & 135.2\\
 & \bmg & \bf{122.0} & \bf{98.6} & \bf{131.2}\\
\midrule
\multirow{4}{*}{RM}
 & \tml & 121.5 & 92.4 & \bf{127.2}\\
 & \bml & 122.9 & 94.0 & 130.4\\
 \cmidrule{2-5}
 & \tmg & \bf{118.6} & \bf{88.9} & 128.8\\
 & \bmg & 129.7 & 96.6 & 135.7\\
\bottomrule
\end{tabular}
\caption{Perplexity comparison including RMN variants with and without temporal matrix (tM) and linear (l) versus gating (g) composition function.}
\label{tb:ppl}
\end{table}

%% file: table-freq.tex
\begin{table}[ht]\centering
\scalebox{0.9}{
\begin{tabular}{@{}l r r r r r r r r r@{}}
\toprule
$d$ & 7 & 8 & 9 & 10 & 11 & 12 &13 &14 &15\\
\midrule
$\mu$ & 54 & 63 & 42 & 67 & 87 & 47 & 67 & 44 & 24\\
\bottomrule
\end{tabular}
}
\caption{Mean frequency ($\mu$) of (\emph{most-attended-word, word-to-predict}) pairs 
grouped by relative distance ($d$).} 
\label{tb:distfreq}
\end{table}

%% file: table-examples.tex
\noindent
\begin{description}\setlength{\itemsep}{.5mm}\setlength{\parskip}{.5mm}%
\item{German:}  findet \emph{statt} (takes \emph{place}),
%einerseits andererseits & one-the-one-hand on-the-other-hand\\
kehrte \emph{zur\"uck} (came \emph{back}),
fragen \emph{antworten} (questions \emph{answers}), 
k\"ampfen \emph{gegen} (fight \emph{against}), 
bleibt \emph{erhalten} (remains \emph{intact}),
verantwortung \emph{\"ubernimmt} (\emph{takes} responsibility);
\item{Italian:} sinistra \emph{destra} (left \emph{right}), latitudine \emph{longitudine}
(latitude \emph{longitude}), collegata \emph{tramite}
(connected \emph{through}), spos\`{o}² \emph{figli}
(got-married \emph{children}), insignito \emph{titolo} (awarded \emph{title}).
\end{description}

\COMMENT{
%\begin{table}[ht]
%\centering
%\begin{adjustbox}{max width=\textwidth}
%\begin{tabular}{\textwidth}{@{} l @{}}
\begin{tabularx}{0.5\textwidth}{@{}l X@{}}
\bf{de} & findet \emph{statt} (takes \emph{place}),
%einerseits andererseits & one-the-one-hand on-the-other-hand\\
kehrte \emph{zur\"uck} (came \emph{back}),
fragen \emph{antworten} (questions \emph{answers}), 
k\"ampfen \emph{gegen} (fight \emph{against}), 
bleibt \emph{erhalten} (remains \emph{intact}),
verantwortung \emph{\"ubernimmt} (\emph{takes} responsibility)\\
\bf{it} & sinistra \emph{destra} (left \emph{right}), latitudine \emph{longitudine}
(latitude \emph{longitude}), collegata \emph{tramite}
(connected \emph{through}), spos\`{o}² \emph{figli}
(got-married \emph{children})\\
\end{tabularx}
%\end{adjustbox}
%\caption{Collocations discovered by RMN. Predicted words are emphasized.}
%\label{tb:examples}
%\end{table}
}

%% file: fig-examples.tex
\begin{figure*}[ht]\small
\setstretch{1}
\begin{mdframed}
\parbox{0.98\textwidth}{
The stage lost a fine \underline{\hspace{2em}}\; , even as science lost an acute reasoner , when he became a specialist in crime\\
a) linguist \phantom{abc} b) hunter \phantom{abc} c) \gold{actor}$^\predrmn$ \phantom{abc} d) estate \phantom{abc} e) horseman$^\predlstm$\\\\
What passion of hatred can it be which leads a man to \underline{\hspace{2em}}\; in such a place at such a time\\
a) \gold{lurk}$^\predrmn$ \phantom{abc} b) dine$^\predlstm$ \phantom{abc} c) luxuriate \phantom{abc} d) grow \phantom{abc}  e) wiggle\\\\
%In my inmost heart I believed that I could \underline{\hspace{2em}}\; where others failed , and now I had the opportunity to test myself\\
%a) smell  \phantom{abc}  b) \gold{succeed}$^\predrmn$ \phantom{abc}  c) lie  \phantom{abc}  d) spell \phantom{abc}  e) forget$^\predlstm$\\\\
%
My heart is \underline{\hspace{2em}}\; already since i have confided my trouble to you\\
a) falling  \phantom{abc}  b) distressed$^\predlstm$ \phantom{abc}  c) soaring  \phantom{abc}  d) \gold{lightened}$^\predrmn$ \phantom{abc}  e) punished \\\\
My morning's work has not been \underline{\hspace{2em}}\; , since it has proved that he has the very strongest motives for standing in the way of anything of the sort\\
a) invisible \phantom{abc} b) neglected$^{\predlstm\predrmn}$ \phantom{abc}  c) overlooked  \phantom{abc} d) \gold{wasted} \phantom{abc}  e) deliberate\\\\
That is his \underline{\hspace{2em}}\; fault , but on the whole he's a good worker\\
a) \gold{main} \phantom{abc} b)  successful \phantom{abc} c)  mother's$^\predrmn$  \phantom{abc} d)  generous \phantom{abc} e) favourite$^\predlstm$
}
\end{mdframed}
\caption{Examples of sentence completion. The correct option is in boldface.
Predictions by the LSTM baseline and by our best RMN model are marked by $^\predlstm$ and $^\predrmn$ respectively. }
\label{fig:sentcomp-examples}
\end{figure*}

%% file: table-acc.tex
\begin{table}[h]\centering
\begin{tabular}{@{}l r r r@{}}
\bf{Model} & $n$ & $d$ & \bf{Accuracy}\\
\toprule
%depRNN & -- & 100 & 52.7\\
LSTM & -- & 256 & 56.0\\
\midrule
\multirow{2}{*}{unidirectional-RM} & 15 & 256 & 64.3\\
 & 15 & 512 & \bf{69.2}\\
\midrule
\multirow{2}{*}{bidirectional-RM}
 & 7 & 256 & 59.6\\
 & 10 & 512 & 67.0\\
\bottomrule
\end{tabular}
\caption{Accuracy on 1,040 test sentences. We use perplexity to choose the best model. Dimension of word embeddings, LSTM hidden states, and gate $g$ parameters are set to $d$.} 
\label{tb:acc}
\end{table}